\documentclass{article}
\usepackage{spconf,amsmath,graphicx,booktabs,multirow,xcolor,amsmath,bbding}
\usepackage[colorlinks=true,urlcolor=blue]{hyperref}

\DeclareMathOperator*{\argmin}{argmin}

\title{BlenDA: Domain Adaptive Object Detection through diffusion-based blending}

\name{Tzuhsuan Huang$^1$$^{\star}$ \qquad Chen-Che Huang$^2$$^{\star}$ \qquad Chung-Hao Ku$^1$ \qquad Jun-Cheng Chen$^1$\thanks{\noindent The first two authors contribute equally to this work. This research is supported by National Science and Technology Council, Taiwan
(R.O.C), under the grant number of NSTC-112-2634-F-002-006 and NSTC-112-2222-E-001-001-MY2, and Academia Sinica under
the grant number of AS-CDA-110-M09.}}

\address{$^1$ Academia Sinica \qquad $^2$ National Taiwan University\\
jason890425@citi.sinica.edu.tw, \{hang6318179, mku666\}@gmail.com, pullpull@citi.sinica.edu.tw}

\begin{document}
\ninept
\maketitle{}
\begin{abstract}
\label{abstract}
\vspace{-5pt}
Unsupervised domain adaptation (UDA) aims to transfer a model learned using labeled data from the source domain to unlabeled data in the target domain. To address the large domain gap issue between the source and target domains, we propose a novel regularization method for domain adaptive object detection, BlenDA, by generating the pseudo samples of the intermediate domains and their corresponding soft domain labels for adaptation training. The intermediate samples are generated by dynamically blending the source images with their corresponding translated images using an off-the-shelf pre-trained text-to-image diffusion model which takes the text label of the target domain as input and has demonstrated superior image-to-image translation quality. Based on experimental results from two adaptation benchmarks, our proposed approach can significantly enhance the performance of the state-of-the-art domain adaptive object detector, Adversarial Query Transformer (AQT). Particularly, in the Cityscapes to Foggy Cityscapes adaptation, we achieve an impressive 53.4\% mAP on the Foggy Cityscapes dataset, surpassing the previous state-of-the-art by 1.5\%. It is worth noting that our proposed method is also applicable to various paradigms of domain adaptive object detection. The code is available at \url{https://github.com/aiiu-lab/BlenDA}
\end{abstract}
\vspace{-5pt}
\section{Introduction}
\label{sec:intro}
\vspace{-5pt}
Unsupervised domain adaptation (UDA) has recently gained significant attention in the field of object detection, with its primary goal being the reduction of cross-domain discrepancies. It enables detectors trained on sufficient labeled source data to effectively generalize to unlabeled data in the target domain.

To bridge the domain gap, recent advancements in unsupervised domain adaptive object detection have prominently featured mean teacher (MT)-based approaches~\cite{MT}. Deng et al.~\cite{UMT} introduce unbiased mean teacher (UMT), incorporating CycleGAN~\cite{CycleGAN} to synthesize images and mitigate domain discrepancies. Another innovative approach is presented by Li et al.~\cite{AT}, who propose an adaptive teacher (AT) utilizing feature-level adversarial learning. This method ensures that features extracted from both source and target domains display similar distributions. Instead of applying adversarial learning, He et al.~\cite{TDD} utilize target-like images along with the proposed target-perceived dual-branch distillation (TDD) framework to enhance the student model by recognizing objects in the target domain through iterative cross-attention. While these methods successfully narrow the domain discrepancy, the quality of the pseudo-labels predicted by the teacher model remains insufficient for domain adaptation.

%To bridge the domain gap, mean teacher (MT)-based approaches~\cite{MT} have recently emerged as a prominent paradigm for enhancing performance in unsupervised domain adaptive object detection. They highlight the presence of model bias when dealing with cross-domain data. Deng et al.~\cite{UMT} introduce unbiased mean teacher (UMT), which uses CycleGAN~\cite{CycleGAN} to synthesize images for reducing domain discrepancies. Li et al.~\cite{AT} propose an adaptive teacher (AT), an approach that employs feature-level adversarial learning. This enables the features extracted from source and target domains to exhibit similar distributions. Rather than applying adversarial learning, He et al.~\cite{TDD} utilize target-like images along with the proposed target-perceived dual-branch distillation (TDD) framework to enhance the student model by recognizing objects in the target domain through iterative cross-attention. While these methods successfully narrow the domain discrepancy, the quality of the pseudo-labels predicted by the teacher model remains insufficient for domain adaptation. %unsupervised learning.
To address this issue, Cao et al.~\cite{CMT} introduce Contrastive Mean Teacher (CMT), utilizing pseudo-labels to extract object-level features and refine them through contrastive learning. Deng et al.~\cite{HT} propose a harmonious teacher (HT) that assesses the quality of pseudo-labels by computing the correlation between object confidence and bounding box location.

\begin{figure}[t]
  \centering
  \includegraphics[width=1.0\linewidth]{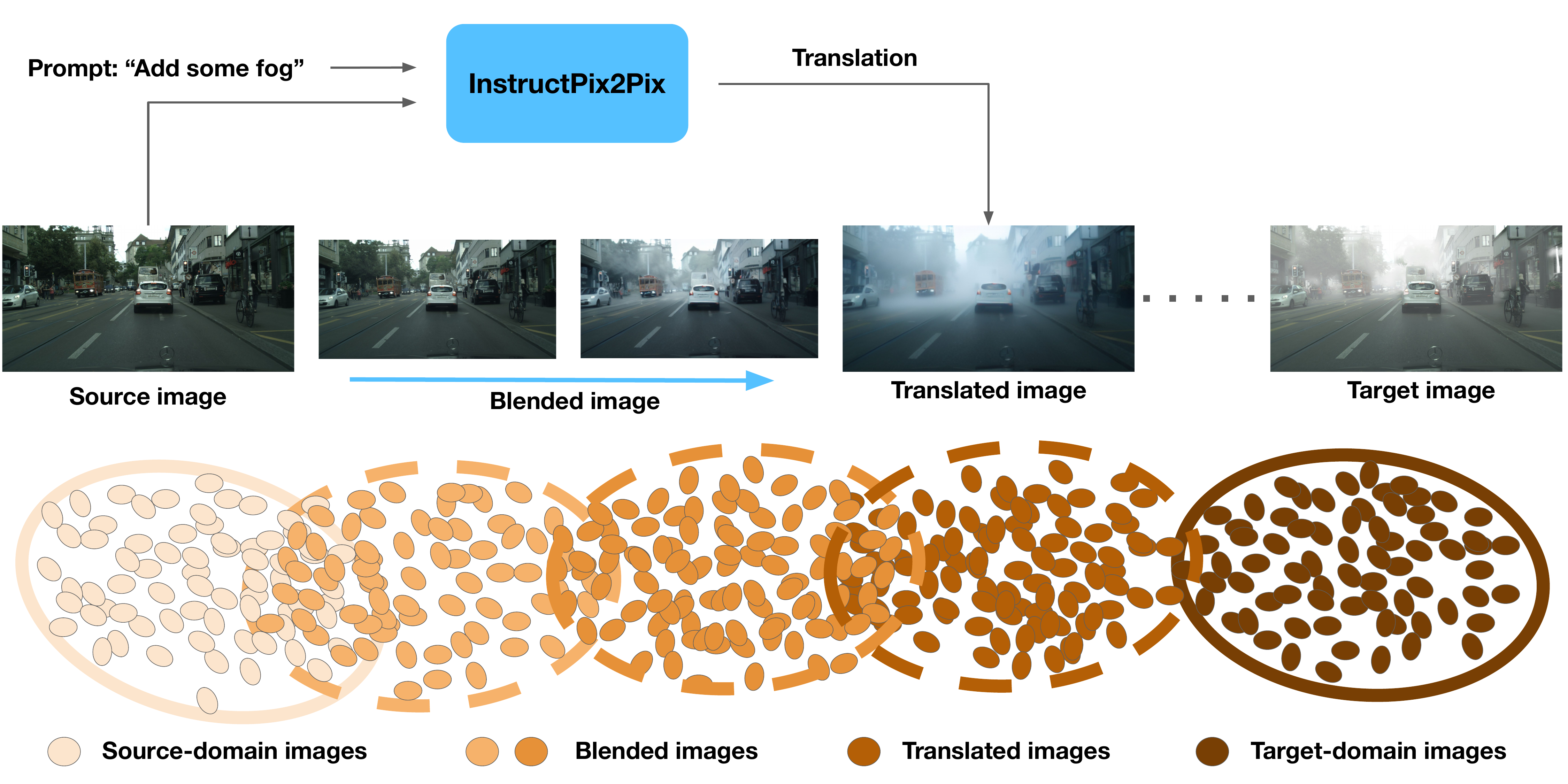}
  \caption{Overview of BlenDA. We utilize a diffusion-based generative model~\cite{Instruction} to generate a target-like translated image with an appropriate prompt and mix it with the source image into a series of blended images for the intermediate domains, which play the role of a bridge between two domains and can be used to train a detector progressively. The direction of the arrow (Blue one) represents an increasing proportion of the translated image used to blend with the source image.}
  \label{fig:teaser}
  \vspace{-10pt}
\end{figure}

Unlike traditional approaches, Mattolin et al.~\cite{ConfMix} propose a method that incorporates a portion of the target domain image with the highest region-level detection confidence into the source image to facilitate domain adaptation. Inspired by data augmentation techniques in image classification tasks~\cite{Mixup}, Vu et al.~\cite{LossMix} train the model by integrating the interpolated image and loss, computed using two distinct labels corresponding to the images used for interpolation.

\begin{figure*}[t]
  \centering
  \includegraphics[width=1.0\linewidth]{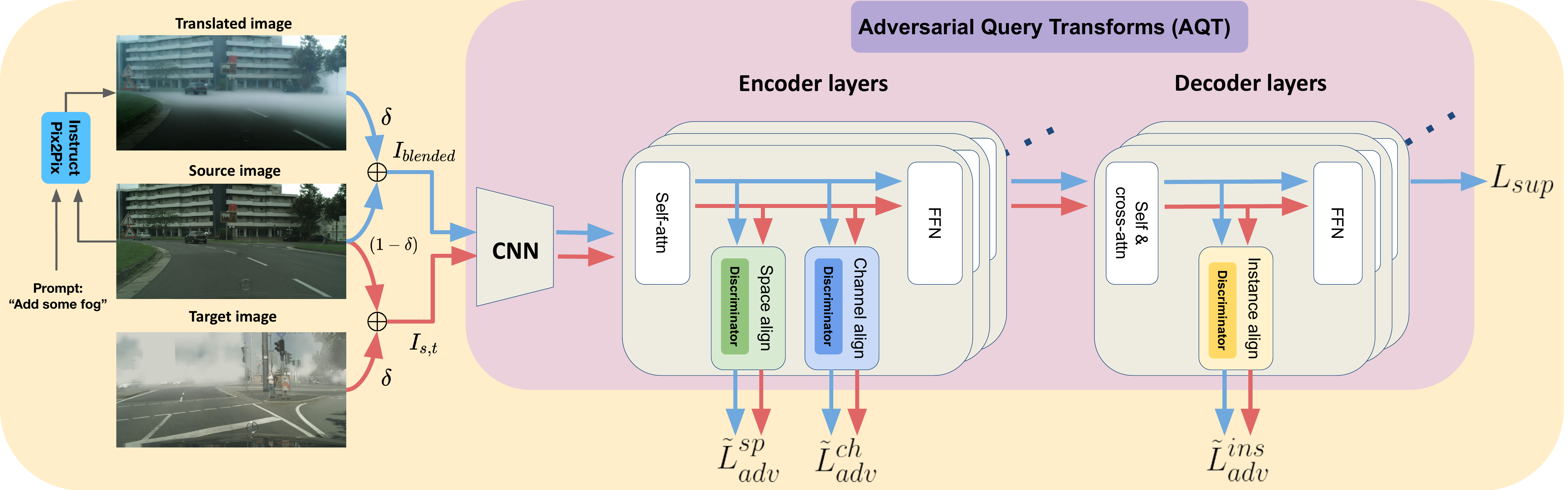}
  \vspace{-15pt}
  \caption{Comprehensive overview of \textbf{BlenDA} on AQT~\cite{AQT}. We first generate translated images by InstructPix2Pix~\cite{Instruction} and compute the $\delta$ using Eq~\eqref{eqn:delta}. Subsequently, we feed the images $I_{blended}$ and $I_{s,t}$, which are mixed according to Eq~\eqref{fix_delta}, into the model. $I_{blended}$ replaces the original source image during the fine-tuning process, using the supervised loss $L_{sup}$ to reduce the domain gap. Moreover, we adjust the original adversarial loss specified in Eq~\eqref{eqn:L_adv} to Eq~\eqref{eqn:L_adv_mixed}. This new adversarial loss enables the discriminators to distinguish different domains in greater detail. Note that the model weights are initialized using the pre-trained weights released by Huang et al.~\cite{AQT} before fine-tuning.}
  \label{fig:overview}
  \vspace{-15pt}
\end{figure*}

\begin{figure}[t]
  \centering
  \includegraphics[width=1.0\linewidth]{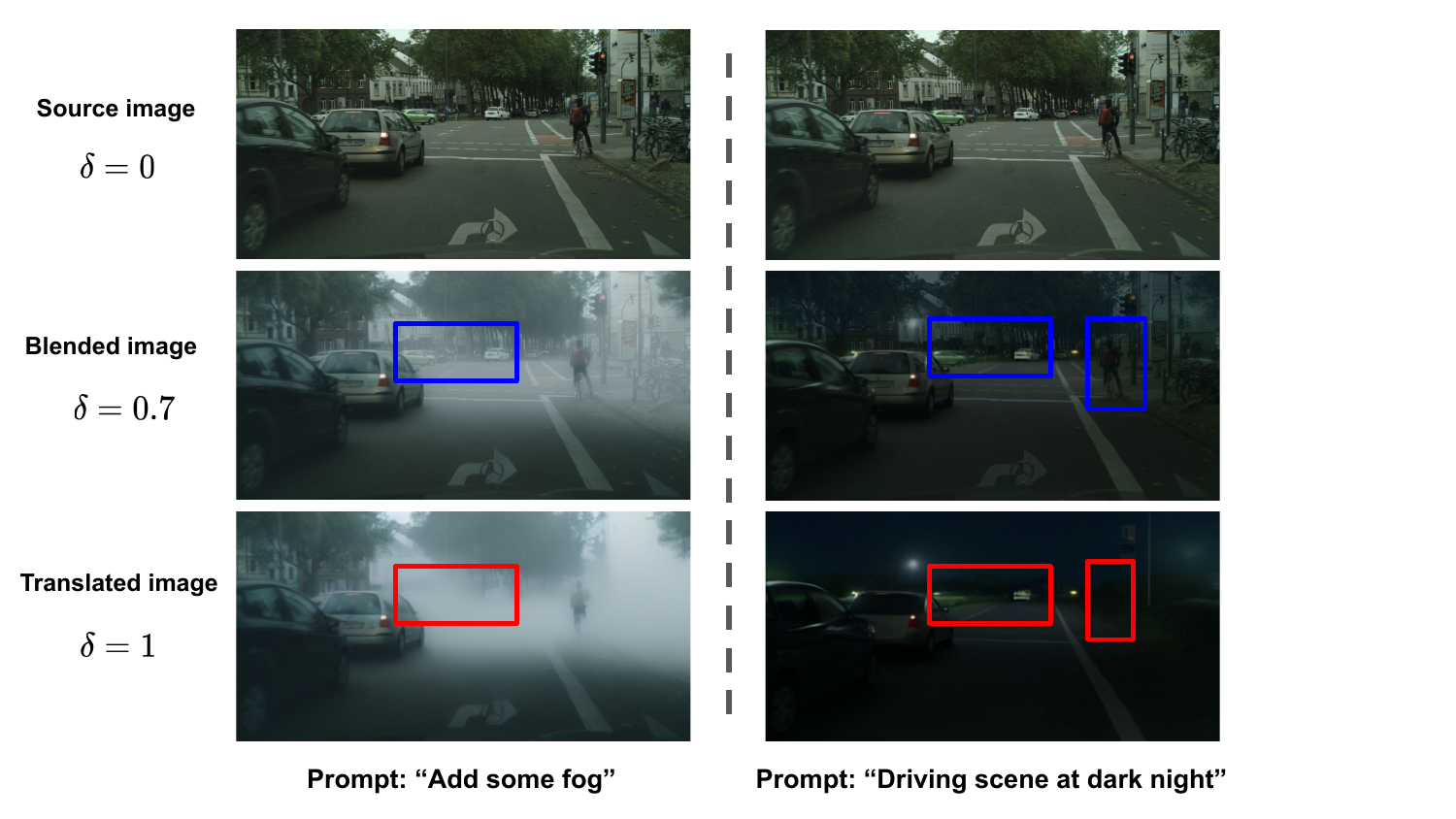}
  \vspace{-20pt}
  \caption{Cityscapes~\cite{Cityscapes} images are fed into InstructPix2Pix~\cite{Instruction} with the prompts to create translated images for both Foggy Cityscapes~\cite{Foggy_Cityscapes} and BDD100K daytime~\cite{BDD_daytime}. Foreground objects are notably absent, as indicated by the red boxes in the translated images. After mixing the source and translated images, foreground objects become visible as outlined by blue boxes in the blended images. $\delta$ represents the proportion used to blend the translated image with the source one.}
  \label{fig:invisible} 
  \vspace{-20pt}
\end{figure}
The methods mentioned above are primarily based on convolution neural network (CNN). Different from CNN-based detection architectures, transformer-based detectors greatly simplifies the two-stage detection pipeline while achieving superior detection performance. Carion et al.~\cite{DETR} propose DETR, which concentrates on token-wise dependencies and eliminates hand-crafted components such as anchor generation and non-maximum suppression, replaced by the Hungarian algorithm. To address slow convergence in DETR, Zhu et al.~\cite{Deformable_DETR} introduce Deformable DETR, which replaces the vanilla attention mechanism in DETR with deformable attention. Recent studies have started to investigate how to leverage transformer-based detectors in domain adaptive object detection, such as Sequence Feature Alignment (SFA)~\cite{SFA}. Based on Deformable DETR, Huang et al.~\cite{AQT} propose an Adversarial Query Transformer (AQT) for domain adaptive object detection, which performs adversarial feature alignment on transformers, with adversarial tokens attending to the feature tokens.
%On the other hand, transformer-based detectors have achieved state-of-the-art performance and dominated object detection. Carion et al.~\cite{DETR} propose DETR which concentrates on token-wise dependencies, and gets rid of hand-crafted components such as anchor generation and non-maximum suppression by the Hungarian algorithm. To mitigate slow convergence in DETR, Zhu et al.~\cite{Deformable_DETR} introduce Deformable DETR, which replaces the vanilla attention mechanism in DETR with the deformable attention. Recent studies~\cite{SFA} suggest that adversarial feature alignment on the convolution neural network (CNN) backbone yields only limited enhancements. Based on Deforamble DETR, Huang et al.~\cite{AQT} propose AQT, which performs adversarial feature alignment on transformers with adversarial tokens attending to the feature tokens.

To improve the domain alignment between the source and target domains, we introduce a novel regularization strategy called BlenDA. The overview is shown in Fig.~\ref{fig:teaser}. During adaptation training, we synthesize high-quality pseudo samples of the intermediate domains by blending source images with their corresponding target-like translated images. These translated images are generated by the off-the-shelf diffusion-based generative model InstructPix2Pix~\cite{Instruction}, which utilizes the text label of the target domain as an instruction for image-to-image translation. Subsequently, the blended images act as a bridge between the source and target domains, serving as labeled data with the soft domain labels and ground-truth annotations for fine-tuning a pre-trained model, which has been trained on data from both the original source and target domains. To further improve the adaptation performance, instead of a fixed mixing weight, we propose dynamically adjusting the mixing weight as the training iteration progresses. Meanwhile, we employ the corresponding soft domain labels to compute adversarial losses, which enhance the capability of discriminators to distinguish between different domains in greater detail. With extensive experiments, the results demonstrate that our approach significantly enhances the model’s performance over the previous state-of-the-art method. 

%%
%As training iteration progresses, the model adapts gradually to fit the blended images and faces the challenge of generalizing to the target domain. Therefore, we allow the mixing weight to adjust dynamically based on the current training iteration. Furthermore, we employ the corresponding soft domain labels for adversarial loss to enhance the discriminator's precision in distinguishing different domains.
%%

\vspace{-5pt}
\section{The Proposed Method}
\label{sec:method}
\vspace{-5pt}
To leverage the superior strength of a pre-trained foundation model for zero-shot text-to-image manipulation, we employ InstructPix2Pix~\cite{Instruction} to produce translated data, establishing a bridge between the source and target domains. In Section \ref{sec:without_mixup}, we commence by pointing out the benefit of blending translated and source images and identifying the problem that may arise when directly using translated images as source data. In Section~\ref{sec:L_adv_new}, we delve into the details of the mixed-domain adversarial loss, enhancing the discriminator's capacity to discern different domains with greater precision. In Section~\ref{sec:overall_training_objective}, we show the overall training objective.

\vspace{-5pt}
\subsection{Sample Generation for Intermediate Domains}
\label{sec:without_mixup}
\vspace{-5pt}
Directly using the translated images as source data for training can lead to transferring inaccurate domain information due to the large domain gap. As indicated by the red boxes in the translated images in Fig.~\ref{fig:invisible}, it is evident that numerous foreground objects are either obscured by fog or engulfed in darkness after translation. Therefore, we first blend the paired source and translated data using a fixed mixing weight $\delta$, as demonstrated in the following equation:
\begin{equation}
\label{fix_delta}
I_{blended}  =\delta \cdot I_{translated} + (1-\delta) \cdot I_{source},
\end{equation}
where $\delta$ is a scalar value, while $I_{blended}$, $I_{translated}$, and $I_{source}$ correspond to the blended image, translated image, and source image, respectively. We use $\delta=0.7$ as an example to demonstrate that combining source and translated images can make foreground objects visible, as indicated by the blue boxes in the blended images shown in Fig.~\ref{fig:invisible}. Subsequently, we utilize the blended image $I_{blended}$ as source data to fine-tune the model, which has been pre-trained on the original source and target data. Intuitively, the presence of the blended image is expected to guide the model towards better alignment with the target domain over time. Nevertheless, as the supervised loss is computed from the blended images and $\delta$ stays constant in the whole training stage, the model may overfit the blended images and cannot generalize well to the target domain. To address this limitation, we propose a method that enables the adaptive adjustment of $\delta$, allowing the model to gradually transfer the acquired domain knowledge to the target domain. Inspired by progressive pseudo-labeling~\cite{ConfMix}, which dynamically adjusts the confidence threshold during non-maximum suppression, we instead use it as a method to blend the source and translated images, in which $\delta$ increases based on the current number of the training iterations:
\vspace{-5pt}
\begin{equation}
\label{eqn:gamma}
\gamma = \frac{\text{current iteration}}{\text{total iterations}},
\end{equation}
\vspace{-7pt}
\begin{equation}
\label{eqn:delta}
\delta = \left( \frac{2}{1+exp(-\alpha \cdot \gamma)} - 1 \right) \cdot \beta,
\end{equation}
where $\gamma$ denotes the training progress, taking on values ranging from 0 to 1. $\alpha$ serves as an adjustment factor; a higher value for $\alpha$ accelerates the removal of source data information. $\beta$ is the upper bound of the mixing weight, ensuring the foreground objects remain visible and are not obscured. See Fig.~\ref{fig:overview}, we generate the translated image and combine it with the paired source image using the dynamic mixing weight $\delta$, as defined in Eq~\eqref{eqn:delta}, and obtain the blended image by Eq~\eqref{fix_delta}. We also blend the target and source images without considering their structural relationship. The way to blend the target and source images is the same as Eq~\eqref{fix_delta}. As illustrated below:
\vspace{-5pt}
\begin{equation}
\label{I_s,t}
I_{s,t}  =\delta \cdot I_{target} + (1-\delta) \cdot I_{source}.
\end{equation}
Subsequently, we use $I_{blended}$ and $I_{s,t}$ as input data to fine-tune the model. Using $I_{blended}$ to fine-tune the model with $L_{sup}$, which is the original detection loss in AQT~\cite{AQT}. Using $I_{blended}$ and $I_{s,t}$ to fine-tune the model with mixed-domain adversarial loss, which will be introduced in Section~\ref{sec:L_adv_new}.

\vspace{-10pt}
\subsection{Mixed-domain Adversarial Loss}
\label{sec:L_adv_new}
\vspace{-5pt}
In this section, we explain how to incorporate the dynamic mixing weight $\delta$ into the adversarial losses of AQT~\cite{AQT}, one of the state-of-the-art domain adaptive object detectors. To perform adversarial learning, gradient reverse layers (GRL)~\cite{GRL} are used to reverse the gradients that pass through the adversarial query tokens during gradient backpropagation. The adversarial losses $L^{sp}_{adv}$, $L^{ch}_{adv}$, and $L^{ins}_{adv}$ in AQT are shown as below:
\begin{equation}
\label{eqn:L_adv}
L^l_{adv} = d \cdot \log D_l (q_l) + (1-d) \cdot \log (1-D_l (q_l)),
\end{equation}
where $l \in \{sp, ch, ins\}$ indicates different levels of feature alignment, with $sp$, $ch$, and $ins$ signifying space-level, channel-level, and instance-level alignment, respectively. $q_l$ is the corresponding adversarial query token, and $D_l$ is the discriminator. Note that $d$ is a domain label, taking value $0$ for the source domain and $1$ for the target domain. To enable the discriminator to distinguish domains in greater detail, instead of the hard domain label, we can use a soft label $\Tilde{d}$ and let $\Tilde{d}=\delta$, as follows:
\begin{equation}
\label{eqn:L_adv_mixed}
\Tilde{L}^l_{adv} = \Tilde{d} \cdot \log D_l(q_l) + (1-\Tilde{d}) \cdot \log (1-D_l(q_l)).
\end{equation}
Note that $\Tilde{d}=\delta$ signifies the mixing weight, indicating the proportion of non-source domain information in the mixing data. For example, it represents the proportion of $I_{target}$ used to blend with the source image for $I_{s,t}$. %it reflects the proportion of $I_{target}$ within $I_{s,t}$. 

\vspace{-5pt}
\subsection{Overall Training Objective}
\label{sec:overall_training_objective}

The total loss is defined as:
\begin{equation}
    \label{total_loss}
    L_{total} = L_{sup} + \lambda_{sp} \tilde{L}^{sp}_{adv} + \lambda_{ch} \tilde{L}^{ch}_{adv} + \lambda_{ins} \tilde{L}^{ins}_{adv},
\end{equation}
where $\lambda_{sp}$, $\lambda_{ch}$, and $\lambda_{ins}$ are the trade-off parameters. The optimization objective for detection transformer $F$ is 
\begin{equation}
    F^* = \argmin_F \max_{\substack{D_{sp}, D_{ch}, \\ D_{ins}}} L_{total}.
\end{equation}

\vspace{-10pt}
\section{Experiments}
\label{experiments}
\vspace{-5pt}

\begin{table*}[t]
\centering
\caption{We present the outcomes and comparative analysis of cross-domain object detection on the validation set of Foggy Cityscapes for the adaptation from \textbf{Cityscapes to Foggy Cityscapes}. The average precision (AP, \%) for all classes is reported.}
\label{tab:city_2_foggy}
\scalebox{1.0}{
\begin{tabular}{cccccccccccc}
\toprule
Method & Detector & Backbone & \textbf{prsn} & \textbf{rider} & \textbf{car} & \textbf{truck} & \textbf{bus} & \textbf{train} & \textbf{motor} & \textbf{bike} & \textbf{mAP} \\
\hline
Source Only & Faster RCNN & R50 & 26.9 & 38.2 & 35.6 & 18.3 & 32.4 & 9.6 & 25.8 & 28.6 & 26.9 \\
TDD~\cite{TDD} & Faster RCNN & R50 & 50.7 & 53.7 & 68.2 & 35.1 & 53.0 & 45.1 & 38.9 & 49.1 & 49.2 \\
CMT~\cite{CMT} & Faster RCNN & V16 & 47.0 & 55.7 & 64.5 & \textbf{39.4} & \textbf{63.2} & \textbf{51.9} & 40.3 & \textbf{53.1} & 51.9 \\
AT~\cite{AT} & Faster RCNN & V16 & 45.5 & 55.1 & 64.2 & 35.0 & 56.3 & 54.3 & 38.5 & 51.9 & 50.1 \\
PT~\cite{PT} & Faster RCNN & V16 & 40.2 & 48.8 & 59.7 & 30.7 & 51.8 & 30.6 & 35.4 & 44.5 & 42.7 \\
\hline
Source Only & FCOS & R50 & 36.9 & 36.3 & 44.1 & 18.6 & 29.3 & 8.4 & 20.3 & 31.9 & 28.2 \\
SIGMA~\cite{SIGMA} & FCOS & R50 & 44.0 & 43.9 & 60.3 & 31.6 & 50.4 & 51.5 & 31.7 & 40.6 & 44.2 \\
SCAN~\cite{SCAN} & FCOS & V16 & 41.7 & 43.9 & 57.3 & 28.7 & 48.6 & 48.7 & 31.0 & 37.3 & 42.1 \\
HT~\cite{HT} & FCOS & V16 & 52.1 & 55.8 & 67.5 & 32.7 & 55.9 & 49.1 & 40.1 & 50.3 & 50.4 \\
OADM~\cite{OADM} & FCOS & V16 & 47.8 & 46.5 & 62.9 & 32.1 & 48.5 & 50.9 & 34.3 & 39.8 & 45.4 \\
\hline
Source Only & Deformable DETR & R50 & 37.7 & 39.1 & 44.2 & 17.2 & 26.8 & 5.8 & 21.6 & 35.5 & 28.5 \\
SFA~\cite{SFA} & Deforamble DETR & R50 & 47.1 & 46.4 & 62.2 & 30.0 & 50.3 & 35.5 & 27.9 & 41.2 & 42.6 \\
AQT~\cite{AQT} & Deformable DETR & R50 & 49.3 & 52.3 & 64.4 & 27.7 & 53.7 & 46.5 & 36.0 & 46.4 & 47.1 \\
AQT(BlenDA) & Deformable DETR & R50 & \textbf{55.0}  & \textbf{57.5} & \textbf{73.6} & 37.5 & 61.6 & 48.0 & \textbf{43.2} & 50.9 & \textbf{53.4} (+6.3) \\
\bottomrule
\end{tabular}
} \vspace{-15pt}
\end{table*}

\begin{table*}[t]
\centering
\caption{We present a performance comparison of AQT, both with and without the integration of BlenDA, on the BDD100K validation set, focusing on the adaptation from \textbf{Cityscapes to BDD100K daytime.}}
\label{tab:city_to_bdd}
\scalebox{1.1}{
\begin{tabular}{lccccccccc}
\toprule
 & & \textbf{prsn} & \textbf{rider} & \textbf{car} & \textbf{truck} & \textbf{bus} & \textbf{motor} & \textbf{bike} & \textbf{mAP} \\
\hline
\textbf{Cityscapes to BDD100K} & AQT & 38.2 & 33.0 & 58.4 & 17.3 & \textbf{18.4} & 16.9 & 23.5 & 29.4 \\
 & AQT(BlenDA) & \textbf{44.5} & \textbf{36.3} & \textbf{64.1} & \textbf{20.0} & 18.1 & \textbf{24.6} & \textbf{26.9} & \textbf{33.5} (+4.1) \\
\bottomrule
\end{tabular}
} \vspace{-16pt}
\end{table*}
\begin{table}[t]
\centering
\caption{Ablation studies of (static/dynamic) $\delta$ and the experiments of different loss settings on \textbf{Cityscapes to Foggy Cityscapes}.}
\label{tab:ablation_studies}
\scalebox{1.0}{
\begin{tabular}{cccc|c}
\\
\toprule
\multicolumn{5}{c}{\textbf{Different $\delta$ Settings}} \\
\midrule
 & \multicolumn{3}{c}{static} & dynamic \\
\midrule
 & $\delta=0.7$ & $\delta=0.9$ & $\delta=1.0$ & $\delta$ in Eq~\eqref{eqn:delta} \\
mAP & 49.1 & 48.0 & \textcolor{red}{32.7} & \textbf{52.8} \\
\midrule
\midrule
\multicolumn{5}{c}{\textbf{W/ or w/o Mixed-domain Adversarial Loss}} \\
\midrule
 & $I_{blended}$ & $I_{target}$ & $I_{s,t}$ & mAP \\
$L_{adv}$ & \Checkmark  & \Checkmark &  & 52.8 \\
$\Tilde{L}_{adv}$ & \Checkmark &  & \Checkmark & 53.4 \\
\bottomrule
\end{tabular}
} \vspace{-15pt}
\end{table}

\subsection{Datasets and Experimental Settings}
\label{sec:dataset}
\vspace{-5pt}
Our experimentation encompasses three standard datasets, including Cityscapes~\cite{Cityscapes}, Foggy Cityscapes~\cite{Foggy_Cityscapes} and BDD100K
~\cite{BDD100K}. We evaluate two adaptation settings: (1) \textbf{Source : Cityscapes} $\rightarrow$ \textbf{Target : Foggy Cityscapes} and (2) \textbf{Source : Cityscapes} $\rightarrow$ \textbf{Target : BDD100K daytime}. The details of the datasets are described below.

% Below is our experimental settings. \\
% \textbf{Source : Cityscapes} $\rightarrow$ \textbf{Target : Foggy Cityscapes} \\
\noindent \textbf{Cityscapes:} A real urban scenes data containing 2,975 images for training and 500 images for validation, including 8 classes (person, rider, car, train, bicycle, motorbike, truck, and bus). \\
\textbf{Foggy Cityscapes:} A synthetic dataset generated from Cityscapes. We take the highest fog density images as the validation set to test the performance of our model.\\
% \textbf{Source : Cityscapes} $\rightarrow$ \textbf{Target : BDD100K daytime} \\
\textbf{BDD100K daytime:} BDD100K is a large-scale driving dataset and the daytime subset is selected as the target domain. The training and validation sets have 70,000 and 10,000 images, respectively. We consider the common 7 categories (person, rider, car, bicycle, motorbike, truck, and bus) following ~\cite{BDD_daytime}. \\

\vspace{-17pt}
\subsection{Implementation Details}
\label{sec:implementation}
\vspace{-5pt}
In our experiments, we retain the original architecture of AQT~\cite{AQT}, while making modifications to the input data and adversarial losses. To create the translated data, in \textbf{Cityscapes to Foggy Cityscapes}, we use InstructPix2Pix~\cite{Instruction} with the prompt ``Add some fog'' and set the text classifier free guidance (cfg) scale = 7.5, image cfg scale = 1.5 and seed = 58,912. In \textbf{Cityscapes to BDD100K daytime}, we use the prompt ``Driving scene at dark night" and set text cfg scale = 9.0, image cfg scale = 1.5 and seed = 981. This translation is applied to all training images in Cityscapes, resulting in a total of 2,975 translated images. For the hyper-parameters, in \textbf{Cityscapes to Foggy Cityscapes}, we set $\alpha=20$, $\beta=1.0$ to control the mixing rate and the mixing upper bound, and $\lambda_{sp}$, $\lambda_{ch}$, and $\lambda_{ins}$ are set to $10^{-1}$. In \textbf{Cityscapes to BDD100K daytime}, we set $\alpha=20$, $\beta=0.5$, $\lambda_{sp}=0.1$, $\lambda_{ch}=10^{-5}$, and $\lambda_{ins}=10^{-4}$. We use AdamW~\cite{AdamW} as the optimizer with the weight decay rate of $10^{-4}$ and the constant learning rate of $2 \times 10^{-5}$ for the entire training process. All experiments are trained for 150 epochs with a batch size of 3 (a source image, a translated image, and a target image), using two NVIDIA RTX A6000 GPUs. 
\vspace{-7pt}
\subsection{Experiment Results Comparison}
\label{sec:results}
\vspace{-5pt}
In this section, we present a performance comparison of AQT using BlenDA alongside other methods on the Cityscapes to Foggy Cityscapes dataset, as shown in Table~\ref{tab:city_2_foggy}. Additionally, we provide a performance analysis of AQT, both with and without the use of BlenDA in Table~\ref{tab:city_2_foggy} and Table~\ref{tab:city_to_bdd}. \\
\textbf{Cityscapes to Foggy Cityscapes:} In Table~\ref{tab:city_2_foggy}, we compare AQT with BlenDA against existing methods built upon Faster RCNN~\cite{Faster_RCNN}, FCOS~\cite{FCOS}, or Deformable DETR~\cite{Deformable_DETR}. ``Source Only'' refers to the baselines that are only trained on source data. ``V16'' and ``R50'' indicate  the backbone architecture used, which are VGG-16~\cite{V16} and ResNet-50~\cite{R50}, respectively. After training with BlenDA, AQT improves by 6.3\% in mAP, outperforming all other methods and surpassing the previous state-of-the-art by 1.5\%. Notably, CMT~\cite{CMT} enhances rare class samples by utilizing target images and their corresponding pseudo labels, whereas BlenDA does not employ such labels. This accounts for the superior performance of rare classes, such as truck, bus, train, and bike, in CMT. \\
\textbf{Cityscapes to BDD100K daytime:} Our method demonstrates a exceptional performance compared to the original AQT, achieving a mAP higher by 4.1\%. Notably, by employing our method, we observe enhancements across nearly all categories, leading to an overall performance surpassing the original AQT baseline. \textbf{Bold text} is employed to signify superior performance.

\vspace{-7pt}
\subsection{Ablation Studies}
\label{sec:ablation_studies}
\vspace{-5pt}
In this section, we analyze the proposed fine-tuning strategy in BlenDA by experiments on \textbf{Cityscapes to Foggy Cityscapes}. \\
\textbf{Different $\delta$ Settings:}
As indicated in Table 3, it is observed that when we set $\delta$ as a hard mixing weight, which disregards the mixing between the source and translated images, the model performs poorly (shown in red). This reflects that even though image translation helps simulate the %incorporate %
domain shifts in the image space, some discriminative details may be lost. Therefore, it is essential to account for mixing between $I_{source}$ and $I_{translated}$ during fine-tuning to achieve %smoothed 
better alignment. 
To further analyze the importance of dynamic mixing weight $\delta$, we use the blended images as source data to fine-tune the model. We set $\delta$ to 0.7 and 0.9 as a static mixing weight to generate $I_{blended}$. Although these blended images with a static mixing weight improve the model's performance, utilizing dynamic $\delta$ can further improve the model as it can progressively control the mixing of source and domain knowledge. See Table~\ref{tab:ablation_studies}, using dynamic mixing weight can achieve performance improvement by 3.7\% compared to the setting of $\delta$ as 0.7, which gives the highest performance for static blending. \\
\noindent \textbf{W/ or w/o Mixed-domain Adversarial Loss:}
To demonstrate that applying the mixed-domain adversarial loss can further enhance the model's performance, we compare the results in two different settings: one uses $I_{blended}$ and $I_{target}$ with the hard domain labels $0$ and $1$, respectively, and the adversarial loss $L_{adv}$; the other utilizes $I_{blended}$ and $I_{s,t}$ with a soft label to compute the mixed-domain adversarial loss $\tilde{L}_{adv}$. As shown in Table~\ref{tab:ablation_studies}, using the mixed-domain adversarial loss $\tilde{L}_{adv}$ for fine-tuning outperforms that without mixed-domain labels by 0.6\% in terms of mAP.

\vspace{-7pt}
\section{Conclusion}
\label{conclusion}
\vspace{-7pt}
In this paper, we introduce BlenDA, a regularization strategy for the training of domain adaptive object detectors that incorporate the mixing of the source and the translated images and their corresponding soft domain labels. In general, the proposed approach is also applicable to various domain adaptive object detection paradigms. 
%With the implementation of BlenDA, the framework can effectively reduce domain bias in source data by generating blended images.
The proposed BlenDA can effectively reduce domain shifts between the source and target domain by generating blended images for the intermediate domains.
We exemplify the application of our training strategy on AQT and evaluate it on two standard benchmarks. The superior results confirm the effectiveness of our methods. The ablation studies further demonstrate that dynamically mixing both images and domain labels significantly improves the model's performance, leading to substantial progress.

% \vspace{-5pt}
% \section{Acknowledgements}
% \label{acknowledgements}
% \vspace{-5pt}
% This research is supported by National Science and Technology Council, Taiwan (R.O.C), under the grant number of NSTC-112-2634-F-002-006 and NSTC-112-2222-E-001-001-MY2, and Academia Sinica under the grant number of AS-CDA-110-M09.
\vfill\pagebreak
\bibliographystyle{IEEEbib}
\bibliography{strings,refs}

\begin{thebibliography}{10}

\bibitem{MT}
Antti Tarvainen and Harri Valpola,
\newblock ``Mean teachers are better role models: Weight-averaged consistency
  targets improve semi-supervised deep learning results,''
\newblock {\em NeurIPS}, 2017.

\bibitem{UMT}
Jinhong Deng, Wen Li, Yuhua Chen, and Lixin Duan,
\newblock ``Unbiased mean teacher for cross-domain object detection,''
\newblock {\em CVPR}, 2021.

\bibitem{CycleGAN}
Jun-Yan Zhu, Taesung Park, Phillip Isola, and Alexei~A. Efros,
\newblock ``Unpaired image-to-image translation using cycle-consistent
  adversarial networks,''
\newblock {\em ICCV}, 2017.

\bibitem{AT}
Yu-Jhe Li, Xiaoliang Dai, Chih-Yao Ma, Yen-Cheng Liu, Kan Chen, Bichen Wu,
  Zijian He, Kris Kitani, and Peter Vajda,
\newblock ``Cross-domain adaptive teacher for object detection,''
\newblock {\em CVPR}, 2022.

\bibitem{TDD}
Mengzhe He, Yali Wang, Jiaxi Wu, Yiru Wang, Hanqing Li, Bo~Li, Weihao Gan, Wei
  Wu, and Yu~Qiao,
\newblock ``Cross domain object detection by target-perceived dual branch
  distillation,''
\newblock {\em CVPR}, 2022.

\bibitem{CMT}
Shengcao Cao, Dhiraj Joshi, Liang-Yan Gui, and Yu-Xiong Wang,
\newblock ``Contrastive mean teacher for domain adaptive object detectors,''
\newblock {\em CVPR}, 2023.

\bibitem{HT}
Jinhong Deng, Dongli Xu, Wen Li, and Lixin Duan,
\newblock ``Harmonious teacher for cross-domain object detection,''
\newblock {\em CVPR}, 2023.

\bibitem{Instruction}
Tim Brooks, Aleksander Holynski, and Alexei~A Efros,
\newblock ``Instructpix2pix: Learning to follow image editing instructions,''
\newblock {\em arXiv preprint arXiv:2211.09800}, 2022.

\bibitem{ConfMix}
Giulio Mattolin, Luca Zanella, Elisa Ricci, and Yiming Wang,
\newblock ``Confmix: Unsupervised domain adaptation for object detection,''
\newblock {\em WACV}, 2023.

\bibitem{Mixup}
Hongyi Zhang, Moustapha Cisse, Yann~N. Dauphin, and David Lopez-Paz,
\newblock ``Mixup: Beyond empirical risk minimization,''
\newblock {\em ICLR}, 2018.

\bibitem{LossMix}
Thanh Vu, Baochen Sun, Bodi Yuan, Alex Ngai, Yueqi Li, and Jan-Michael Frahm,
\newblock ``Lossmix: Simplify and generalize mixup for object detection and
  beyond,''
\newblock {\em arXiv preprint arXiv:2303.10343}, 2023.

\bibitem{AQT}
Wei-Jie Huang, Yu-Lin Lu, Shih-Yao Lin, Yusheng Xie, and Yen-Yu Lin,
\newblock ``Aqt: Adversarial query transformers for domain adaptive object
  detection,''
\newblock {\em IJCAI-ECAI}, 2022.

\bibitem{Cityscapes}
Marius Cordts, Mohamed Omran, Sebastian Ramos, Timo Rehfeld, Markus Enzweiler,
  Rodrigo Benenson, Uwe Franke, Stefan Roth, and Bernt Schiele,
\newblock ``The cityscapes dataset for semantic urban scene understanding,''
\newblock {\em CVPR}, 2016.

\bibitem{Foggy_Cityscapes}
Christos Sakaridis, Dengxin Dai, and Luc~Van Gool,
\newblock ``Semantic foggy scene understanding with synthetic data,''
\newblock {\em IJCV}, 2018.

\bibitem{BDD_daytime}
Minghao Xu, Hang Wang, Bingbing Ni, Qi~Tian, and Wenjun Zhang,
\newblock ``Cross-domain detection via graph-induced prototype alignment,''
\newblock {\em CVPR}, 2020.

\bibitem{DETR}
Nicolas Carion, Francisco Massa, Gabriel Synnaeve, Nicolas Usunier, Alexander
  Kirillov, and Sergey Zagoruyko,
\newblock ``End-to-end object detection with transformers,''
\newblock {\em ECCV}, 2020.

\bibitem{Deformable_DETR}
Xizhou Zhu, Weijie Su, Lewei Lu, Bin Li, Xiaogang Wang, and Jifeng Dai,
\newblock ``Deformable detr: Deformable transformers for end-to-end object
  detection,''
\newblock {\em ICLR}, 2021.

\bibitem{SFA}
Wen Wang, Yang Cao, Jing Zhang, Fengxiang He, Zheng-Jun Zha, Yonggang Wen, and
  Dacheng Tao,
\newblock ``Exploring sequence feature alignment for domain adaptive detection
  transformers,''
\newblock {\em ACM MM}, 2021.

\bibitem{GRL}
Yaroslav Ganin and Victor Lempitsky,
\newblock ``Unsupervised domain adaptation by backpropagation,''
\newblock {\em ICML}, 2015.

\bibitem{PT}
Meilin Chen, Weijie Chen, Shicai Yang, Jie Song, Xinchao Wang, Lei Zhang,
  Yunfeng Yan, Donglian Qi, Yueting Zhuang, Di~Xie, and Shiliang Pu,
\newblock ``Learning domain adaptive object detection with probabilistic
  teacher,''
\newblock {\em ICML}, 2022.

\bibitem{SIGMA}
Wuyang Li, Xinyu Liu, and Yixuan Yuan,
\newblock ``Sigma: Semantic-complete graph matching for domain adaptive object
  detection,''
\newblock {\em CVPR}, 2022.

\bibitem{SCAN}
Wuyang Li, Xinyu Liu, Xiwen Yao, and Yixuan Yuan,
\newblock ``Scan: Cross domain object detection with semantic conditioned
  adaptation,''
\newblock {\em AAAI}, 2022.

\bibitem{OADM}
Jayeon Yoo, Inseop Chung, and Nojun Kwak,
\newblock ``Unsupervised domain adaptation for one-stage object detector using
  offsets to bounding box,''
\newblock {\em ECCV}, 2022.

\bibitem{BDD100K}
Fisher Yu, Haofeng Chen, Xin Wang, Wenqi Xian, Yingying Chen, Fangchen Liu,
  Vashisht Madhavan, and Trevor Darrell,
\newblock ``Bdd100k: A diverse driving dataset for heterogeneous multitask
  learning,''
\newblock {\em CVPR}, 2020.

\bibitem{AdamW}
Ilya Loshchilov and Frank Hutter,
\newblock ``Decoupled weight decay regularization,''
\newblock {\em ICLR}, 2019.

\bibitem{Faster_RCNN}
Shaoqing Ren, Kaiming He, Ross Girshick, and Jian Sun,
\newblock ``Faster r-cnn: Towards real-time object detection with region
  proposal networks,''
\newblock {\em NIPS}, 2015.

\bibitem{FCOS}
Zhi Tian, Chunhua Shen, Hao Chen, and Tong He,
\newblock ``Fcos: Fully convolutional one-stage object detection,''
\newblock {\em ICCV}, 2019.

\bibitem{V16}
Karen Simonyan and Andrew Zisserman,
\newblock ``Very deep convolutional networks for large-scale image
  recognition,''
\newblock {\em ICLR}, 2015.

\bibitem{R50}
Kaiming He, Xiangyu Zhang, Shaoqing Ren, and Jian Sun,
\newblock ``Deep residual learning for image recognition,''
\newblock {\em CVPR}, 2016.

\end{thebibliography}
\end{document}